# Comprehensive Multimodal Segmentation in Medical Imaging: Combining YOLOv8 with SAM and HQ-SAM Models

Sumit Pandey
Department of Computer Science
University of Copenhagen
supa@di.ku.dk

Kuan-Fu Chen
Department of Emergency Medicine
Chang Gung Memorial Hospital
drkfchen@gmail.com

Erik B. Dam
Department of Computer Science
University of Copenhagen
erikdam@di.ku.dk

## Abstract

*This paper introduces a comprehensive approach for segmenting regions of interest (ROI) in diverse medical imaging datasets, encompassing ultrasound, CT scans, and X-ray images. The proposed method harnesses the capabilities of the YOLOv8 model for approximate boundary box detection across modalities, alongside the Segment Anything Model (SAM) and High Quality (HQ) SAM for fully automatic and precise segmentation. To generate boundary boxes, the YOLOv8 model was trained using a limited set of 100 images and masks from each modality.*

*The results obtained from our approach are extensively computed and analyzed, demonstrating its effectiveness and potential in medical image analysis. Various evaluation metrics, including precision, recall, F1 score, and Dice Score, were employed to quantify the accuracy of the segmentation results. A comparative analysis was conducted to assess the individual and combined performance of the YOLOv8, YOLOv8+SAM, and YOLOv8+HQ-SAM models.*

*The results indicate that the SAM model performs better than the other two models, exhibiting higher segmentation accuracy and overall performance. While HQ-SAM offers potential advantages, its incremental gains over the standard SAM model may not justify the additional computational cost. The YOLOv8+SAM model shows promise for enhancing medical image segmentation and its clinical implications.*

## 1. Introduction

Medical image segmentation plays a crucial role in a wide range of medical image analysis tasks and serves as a vital component in computer-aided diagnosis and pathology research. The field of medical image segmentation has witnessed significant advancements, particularly with the emergence of convolutional neural networks (CNNs) in computer vision. These CNNs, specifically Encoder-Decoder based architectures, have demonstrated remarkable success in various medical imaging applications such as brain Magnetic Resonance Imaging (MRI) [10], multi-organ segmentation, and cardiac ventricle analysis [13] [5]. Their capacity for end-to-end image semantic segmentation has led to the development of notable variants like U-Net [16], 3D U-Net [4], tailored to segmenting images and volumes from different medical imaging modalities.

Despite the success of CNN-based methods, they encountered challenges in handling long-distance dependencies between image elements due to limited convolution kernel size. To overcome these challenges and effectively incorporate global contextual information, the Transformer [18] was introduced. Initially designed for sequence-to-sequence prediction tasks, the Transformer architecture revolutionized the field of natural language processing. Subsequently, it found application in medical image segmentation, leading to the creation of State of the art (SOTA) models like Attention U-Net [14], along with its variations such as Multi-res-attention UNet [17] and Attention Res-UNet [12]. Unlike traditional CNN-based methods, the Transformer architecture adopts a fully attention-based encoder-decoder structure, operating on one-dimensional sequences, which endows it with robust modeling capabilities for capturing global context information. Furthermore, the Transformer's potential is further harnessed through large-scale pre-training, making it adaptable to various downstream tasks in the medical image analysis domain.

The Transformer has emerged as a powerful tool for medical image segmentation, but its emphasis on global context information often comes at the cost of capturing fine local details. This limitation affects the accuracy in distinguishing between background and target regions. To address this issue and build upon the Transformer's strengths, researchers have integrated Convolutional Neural Networks (CNNs) with Transformer architectures. One promising approach is TransUNet [1], which cleverly combines CNNs to extract local features and Transformer for global context modeling. By incorporating a self-attention mechanism, TransUNet successfully retains the resolution of local features, resulting in significant improvements in image segmentation accuracy. Despite these advancements, TransUNet represents only a preliminary integration of CNNs and Transformer, leaving ample room for practical enhancements and refinements.

A crucial aspect of semantic segmentation lies in the extraction and fusion of low-dimensional image texture features, including structural and statistical features. These features significantly impact the segmentation performance. For example, [3] introduced DeepLabv3+, which enhances segmentation by incorporating an encoder into the DeepLabv3 model [2]. This modification facilitates the extraction and fusion of both shallow and deep image features. Similarly, [9] addressed the importance of edge features by introducing an edge preservation module, resulting in a notable overall boost in semantic segmentation performance. While these methods have highlighted the significance of low-dimensional features, few existing solutions have delved into the analysis of low-dimensional statistical features for grasping global image characteristics. By exploring the potential of these features, it may be possible to further improve the performance of medical image segmentation models and achieve even more accurate and reliable results.

In this paper, we propose a comprehensive approach for ROI segmentation in diverse medical imaging datasets by combining the YOLOv8 [6] model with the Segment Anything Model (SAM) [8] and the High Quality (HQ) SAM [7]. We aim to harness the strengths of both CNN-based models and the Transformer architecture, facilitating accurate and efficient ROI segmentation with improved global context modeling and detailed local feature preservation. The integration of YOLOv8 with SAM, and HQ-SAM offers a promising solution to address the challenges in medical image analysis and enhance the accuracy and efficiency of ROI segmentation in various medical imaging modalities. The experimental results demonstrate the effectiveness of our proposed approach, emphasizing its potential to advance medical image analysis tasks and improve patient care through automated and reliable ROI segmentation.

## 2. Methodology

Our study follows a three-part methodology. The first part involves data collection. In the second part, we train the YOLOv8 model using a limited dataset consisting of 100 images and masks to create approximate boundary boxes around the objects of interest. Finally, in the third part, we integrate YOLOv8 with SAM and HQ-SAM models to generate segmentation masks for ROI.

### 2.1. Dataset Selection and Preprocessing

The first part of our study was Data collection and preprocessing, this study made use of three distinct datasets representing different modalities. The first dataset, acquired from Chang Gung Memorial Hospital, is called the Ultrasound Short-Axis Aorta Segmentation Dataset. It comprises 200 images of patients, along with their corresponding masks for segmentation. The second dataset utilized in the study is the Lung CT Scan Segmentation Dataset, which is an open-source dataset obtained from Kaggle [11]. This dataset contains 267 2D images, each accompanied by its respective masks for segmentation. Lastly, the third dataset employed in the study is also sourced from Kaggle and is known as the Lung X-ray Segmentation Dataset. This dataset contains 705 2D images, and similar to the others, it includes masks for segmentation [15]. By combining these three datasets, the study aims to explore and analyze various modalities in the context of segmentation tasks.

After data collection we performed preprocessing, this step includes cross-reference the images and masks, resizing, normalization, and augmentation to enhance the dataset's variability and generalizability.

### 2.2. Training YOLOv8 for ROI Detection and Segmentation

The second part of our study focuses on training the YOLOv8 model using a meticulously curated multimodal medical image dataset. This dataset is a combination of three different datasets, each representing a distinct modality. The main objective of the YOLOv8 model is to accurately detect approximate boundary boxes around the Regions of Interest (ROI) present in these medical images.

After conducting few experiments with different numbers of image-mask pairs, we have come to the conclusion that training the YOLOv8 model on 100+ image-mask pairs yields satisfactory results in generating boundary boxes for the ROIs. As a result, we have decided to train the YOLOv8 model using only 100 randomly selected image-mask pairs from our datasets.

The reason for aiming at approximate boundary boxes is that the SAM model's dice score remains relatively consistent even when the boundary boxes vary by a small margin, specifically 5-10 pixels. To validate this claim, we conducted an experiment and plotted the box plot of dice scores obtained from the SAM model. we varied pixel distance for boundary boxes from perfectly hand-curated boundary boxes (DS0) to DS5 (increased by 5 pixels from all four corners), DS10 (increased by 10 pixels from all four corners), and so on. The results in Figure 1 indicate that the dice score indeed remains relatively consistent when the pixel distance is increased by 5-10 pixels. This finding suggests that having approximate boundary boxes generated by the YOLOv8 model is sufficient for the subsequent SAM model's performance.

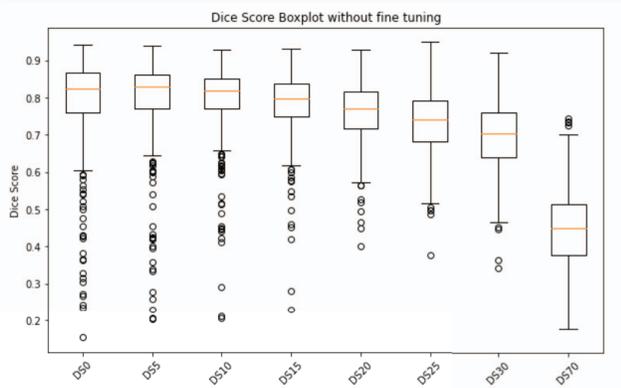

Figure 1. This figure shows the dice scores' (SAM model on ultrasound images) box plots when bounding boxes (manually drawn) are perfect fit to ROI (DS0), when bounding boxes are 5 pixels are bigger (DS5), when bounding boxes are 10 pixels are bigger (DS10) and so on.

It is crucial to clarify that in this work, the YOLOv8 model is solely responsible for generating boundary boxes and not segmentation masks. This is because the limited dataset of images and masks would not be sufficient to generate accurate segmentation masks.

### 2.3. Generating Boundary Boxes and Utilizing SAM and HQ-SAM for Segmentation

The third and final part of our study is to combine YOLOV8 with SAM models (SAM and HQ-SAM). SAM is a versatile and potent architecture designed for real-world segmentation tasks. Its strengths lie in supporting flexible prompts, real-time mask computation, and its ability to handle ambiguity. Utilizing image, prompt, and lightweight mask encoders, SAM accurately predicts segmentation masks, even generating multiple masks for a single prompt to adapt to ambiguous scenarios. Its effectiveness is further enhanced by being trained on a diverse set of masks using the innovative "data engine" strategy, resulting in high-quality and real-time mask predictions [8].

Building upon SAM's foundation, HQ-SAM is an advanced segmentation model that incorporates a learnable High-Quality Output Token. This augmentation enables precise object segmentation without significantly increasing parameters or computational complexity, while still maintaining promptability and zero-shot generalizability [7].

Figure 2 depicts the process of generating segmentation predicted boundary boxes using the YOLOv8 model. These boundary boxes serve as essential inputs to both SAM and HQ-SAM models, acting as regions of interest (ROIs). By directing the attention of subsequent models to relevant areas within the image, these boundary boxes aid in achieving accurate segmentation results.

The decision to employ SAM was based on its robustness and effectiveness in handling ambiguity, particularly concerning boundary boxes in this case. Integrating YOLOv8 with SAM and HQ-SAM models aims to achieve superior segmentation results compared to using segmentation masks directly from YOLOv8 . By combining YOLOv8's approximate boundary boxes with the spatial attention mechanisms of SAM and HQ-SAM, this approach ensures better localization and segmentation of regions of interest in medical images.

### 3. Results

Figure 3 displays visual results of random images, along with their ground truth labels and the predicted masks by SAM (HQ-SAM, SAM) and YOLOv8 models. Upon visual comparison, both SAM models (HQ-SAM and SAM) demonstrate superior performance in all three modalities, as their predicted masks are significantly better than YOLOv8's. However, it's important to note that YOLOv8 is solely used for generating prompts for the SAM models and it was expected that YOLOv8 will perform bad on segmentation as compare to SAM models. The inclusion of YOLOv8's segmentation results in the paper aims to demonstrate its inferior performance in predicting segmentation masks, while also showcasing how the prompts (boundary boxes) derived from YOLOv8 enable full automation in the SAM models. To delve deeper into the results, a computational analysis of YOLOv8+SAM and YOLOv8+HQ-SAM was conducted.

In the X-ray dataset, the SAM model exhibited strong segmentation capabilities with a mean Dice Score of 0.9012, Precision of 0.8747, Recall of 0.9419, and F1-score of 0.9012. The HQ-SAM model, designed to provide higher quality segmentation, achieved slightly lower results, with a mean Dice Score of 0.8902, Precision of 0.8434, Recall of 0.9560, and F1-score of 0.8902. However, both SAM and HQ-SAM models significantly outperformed the YOLOv8 model, which struggled with a mean Dice Score of only 0.1938, Precision of 0.173, Recall of 0.241, and F1-score of 0.1938 on the same dataset (as shown in table 1 and Figure 4).

Moving on to the Short-axis Aorta Ultrasound dataset, the SAM model once again demonstrated its efficacy with a mean Dice Score of 0.769, Precision of 0.836, Recall of

2594

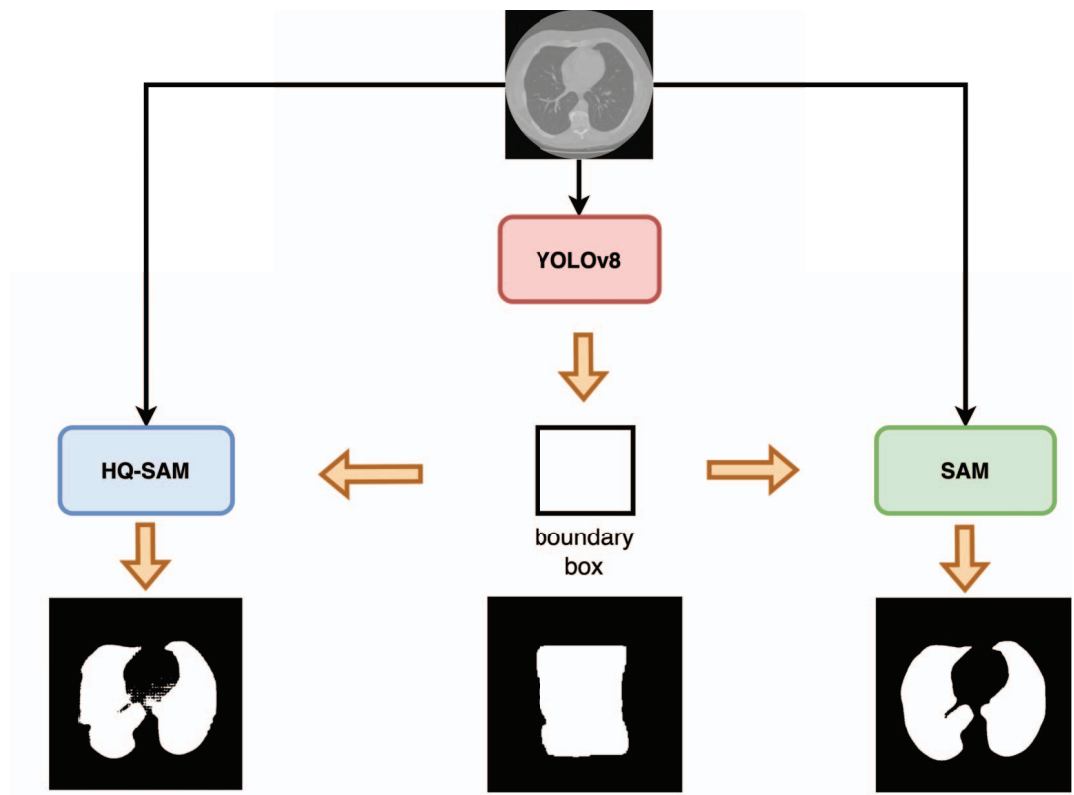

Figure 2. We use YOLOv8 to generate boundary boxes as prompt on Lung CT Scan images. These prompts are then fed into SAM and HQ-SAM models, which produce segmentation masks for the Regions of Interest (ROI).

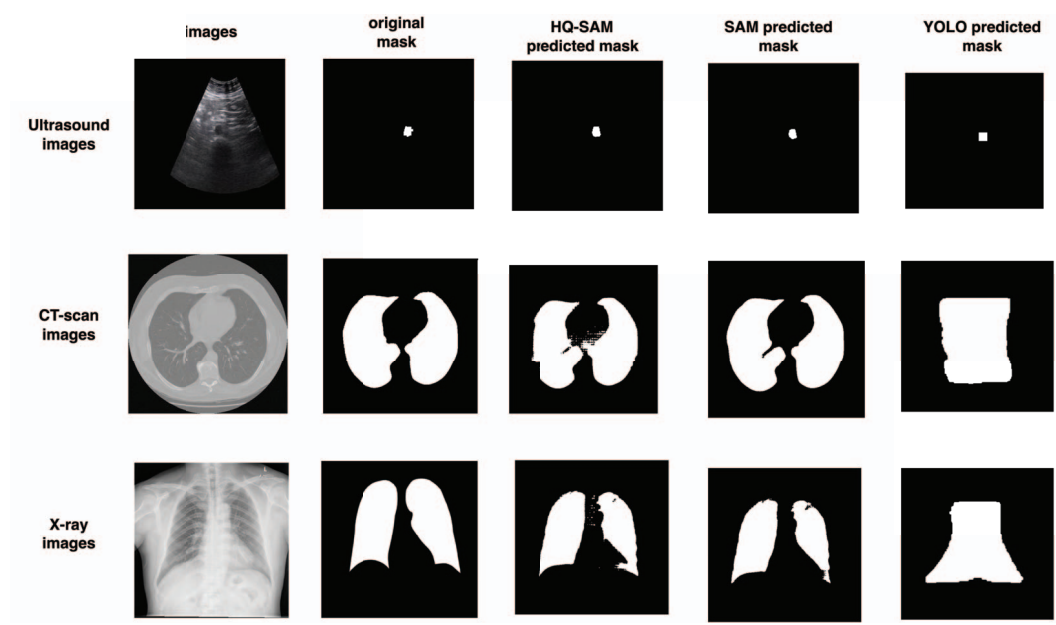

Figure 3. The image displays random medical images with their ground truth labels and predicted masks from SAM (HQ-SAM, SAM) and YOLOv8 models.

0.732, and F1-score of 0.769. The HQ-SAM model's performance was similar to that of SAM, achieving a mean

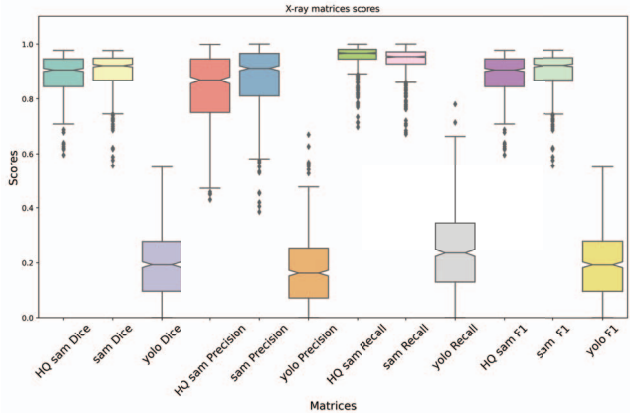

Figure 4. Box plot Results matrices for X-ray Dataset.

| Stats | HQ sam Dice Score | HQ sam Precision | HQ sam Recall | HQ sam F1-score |
|---|---|---|---|---|
| Mean | 0.8902 | 0.8434 | 0.9560 | 0.8902 |
| Median | 0.9052 | 0.868 | 0.966 | 0.9052 |
| Std. | 0.0669 | 0.1166 | 0.0407 | 0.066 |
| Stats | SAM Dice Score | SAM Precision | SAM Recall | SAM F1-score |
| Mean | 0.9012 | 0.8747 | 0.9419 | 0.9012 |
| Median | 0.9210 | 0.9112 | 0.953 | 0.9210 |
| Std. | 0.0633 | 0.1120 | 0.0474 | 0.0633 |
| Stats | YOLO Dice Score | YOLO Precision | YOLO Recall | YOLO F1-score |
| Mean | 0.1938 | 0.173 | 0.241 | 0.1938 |
| Median | 0.193 | 0.162 | 0.2363 | 0.1932 |
| Std. | 0.123 | 0.1227 | 0.149 | 0.123 |

Table 1. Results matrices for X-ray Dataset.

Dice Score of 0.7722, Precision of 0.834, Recall of 0.7392, and F1-score of 0.772. While the YOLOv8 model improved compared to its X-ray results, it still fell short, yielding a mean Dice Score of 0.5064, Precision of 0.8775, Recall of 0.4254, and F1-score of 0.5064 (as shown in table 2 and Figure 5).

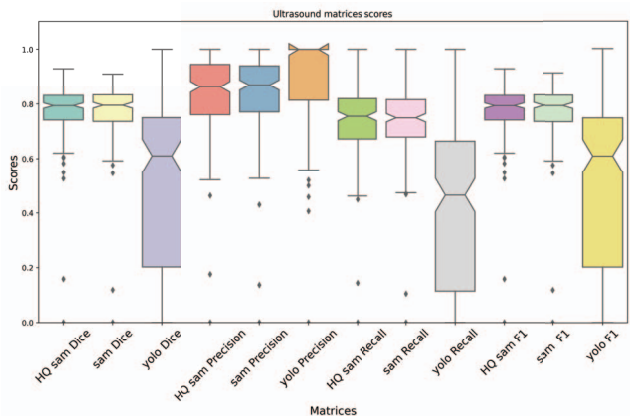

Figure 5. Box plot results matrices for Short-axis Aorta Ultrasound dataset.

For the lung CT scan segmentation dataset, both SAM and HQ-SAM models showcased their effectiveness. The SAM model obtained a mean Dice Score of 0.8799, Precision of 0.836, Recall of 0.948, and F1-score of 0.879.

| Stats | HQ sam Dice Score | HQ sam Precision | HQ sam Recall | HQ sam F1-score |
|---|---|---|---|---|
| Mean | 0.7722 | 0.834 | 0.7392 | 0.772 |
| Median | 0.7946 | 0.864 | 0.755 | 0.7946 |
| Std. | 0.1157 | 0.1533 | 0.1438 | 0.115 |
| Stats | SAM Dice Score | SAM Precision | SAM Recall | SAM F1-score |
| Mean | 0.769 | 0.836 | 0.732 | 0.769 |
| Median | 0.795 | 0.870 | 0.75 | 0.795 |
| Std. | 0.116 | 0.1529 | 0.143 | 0.116 |
| Stats | YOLO Dice Score | YOLO Precision | YOLO Recall | YOLO F1-score |
| Mean | 0.5064 | 0.8775 | 0.4254 | 0.5064 |
| Median | 0.6100 | 1 | 0.4651 | 0.6100 |
| Std. | 0.2927 | 0.2232 | 0.3017 | 0.2927 |

Table 2. Results matrices for Short-axis Aorta Ultrasound Dataset.

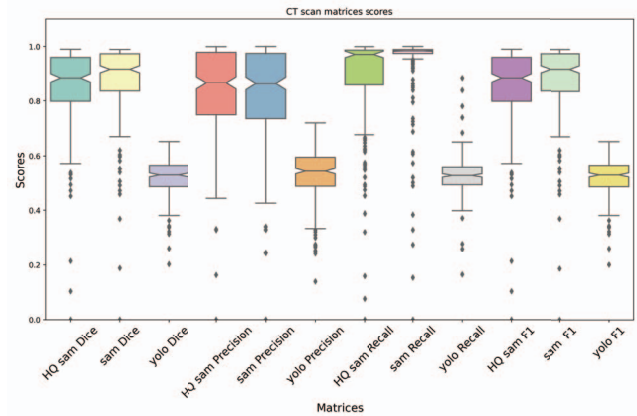

Figure 6. Box plot results matrices for lung CT scan segmentation dataset.

The HQ-SAM model also performed well with a mean Dice Score of 0.8554, Precision of 0.8429, Recall of 0.8903, and F1-score of 0.855. Once again, the YOLOv8 model lagged behind, achieving a mean Dice Score of 0.520, Precision of 0.5291, Recall of 0.525, and F1-score of 0.520 (as shown in table 3 and Figure 6).

| Stats | HQ sam Dice Score | HQ sam Precision | HQ sam Recall | HQ sam F1-score |
|---|---|---|---|---|
| Mean | 0.8554 | 0.8429 | 0.8903 | 0.855 |
| Median | 0.8839 | 0.8659 | 0.970 | 0.883 |
| Std. | 0.139 | 0.1557 | 0.1630 | 0.139 |
| Stats | SAM Dice Score | SAM Precision | SAM Recall | SAM F1-score |
| Mean | 0.8799 | 0.836 | 0.948 | 0.879 |
| Median | 0.9160 | 0.86 | 0.983 | 0.9160 |
| Std. | 0.131 | 0.1582 | 0.1292 | 0.131 |
| Stats | YOLO Dice Score | YOLO Precision | YOLO Recall | YOLO F1-score |
| Mean | 0.520 | 0.5291 | 0.525 | 0.520 |
| Median | 0.5301 | 0.5446 | 0.52 | 0.530 |
| Std. | 0.0652 6 | 0.095 | 0.0674 | 0.066 |

Table 3. Results matrices for lung CT scan segmentation dataset.

## 4. Discussion

The findings presented in this paper provide valuable insights into the performance of three different models, SAM, HQ-SAM, and YOLOv8, on various medical imaging datasets. Medical image segmentation is a crucial task in healthcare, enabling accurate identification and delineation of anatomical structures, tumors, and other abnormalities. The results of this study shed light on the strengths and limitations of each model and offer important considerations for researchers and practitioners working in the field of medical image analysis.

The SAM model consistently outperformed the other two models, HQ-SAM and YOLOv8, across most datasets. Its mean Dice Score, Precision, Recall, and F1-score consistently exhibited higher values, indicating superior segmentation accuracy and overall performance. The robustness and effectiveness of the SAM model can be attributed to its underlying architecture, which likely includes advanced convolutional neural networks (CNNs) and attention mechanisms. These components enable the model to effectively learn and represent complex patterns in medical images, leading to more accurate segmentations.

The HQ-SAM model was developed with the intention of providing higher quality segmentation results. However, the results indicate that the improvements achieved with HQ-SAM over the standard SAM model were not substantial in most cases. While the HQ-SAM model still demonstrated competitive performance, the marginal gains in accuracy and precision may not justify the additional computational cost required for its implementation. This suggests that practitioners and researchers should carefully weigh the benefits of using HQ-SAM against its increased computational demands, especially in real-time or resource-constrained environments.

The YOLOv8 model consistently struggled to produce accurate segmentations across all datasets. Its mean Dice Score, Precision, Recall, and F1-score were significantly lower compared to both SAM and HQ-SAM models. YOLOv8's inferior performance may be attributed to several factors, including trained on limited dataset, its architecture's suitability for medical imaging tasks and its limited ability to handle complex and diverse anatomical structures. But in our study the main goal of YOLOv8 model is to predicted approx. boundary boxes around ROI and by looking at the performances of SAM and HQ-SAM on all 3 medical modalities, we can conclude that this technique worked well.

The results underscore the importance of carefully selecting the appropriate model based on the specific requirements of the medical imaging task and the characteristics of the dataset. While the SAM model emerged as the top performer overall, its advantages should be weighed against the specific application and resource constraints. For instance, if real-time processing is a crucial requirement, the standard SAM model might be a more practical choice than HQ-SAM.

Interestingly, the performance of each model varied across datasets. While the SAM model achieved outstanding results in both X-ray and lung CT scan segmentation, its performance on the Short-axis Aorta Ultrasound dataset was slightly lower. This emphasizes the need to evaluate models on multiple datasets to gain a comprehensive understanding of their strengths and limitations. Additionally, dataset-specific characteristics, such as image quality, resolution, and class imbalances, can significantly influence model performance and should be carefully considered during the model selection process.

The results suggest that there is still room for improvement in medical image segmentation models. Further research could focus on enhancing the SAM model or exploring new architectures and attention mechanisms to push the boundaries of segmentation accuracy. Additionally, model ensembles or combining the strengths of multiple models could be investigated to potentially achieve even higher segmentation performance.

Accurate medical image segmentation is of paramount importance in clinical practice for disease diagnosis, treatment planning, and monitoring patient progress. The superior performance of the SAM model has promising clinical implications, as it can aid healthcare professionals in making more precise and informed decisions. Moreover, accurate segmentation can lead to improved automated analysis, reducing the burden on radiologists and enabling faster diagnosis and treatment.

It is essential to acknowledge the limitations of this study. The evaluation was limited to three specific models, and other state-of-the-art models might have been excluded. Additionally, the datasets used in the study may not fully represent the diversity of medical imaging challenges, and performance on other datasets might differ. Future studies should incorporate more diverse datasets and consider the transferability of the models to different medical imaging modalities.

## 5. Conclusion

In this paper, we conducted a comprehensive evaluation of three different models, namely, SAM, HQ-SAM, and YOLOv8, for medical image segmentation across multiple datasets. Medical image segmentation is a critical task in healthcare, enabling precise identification and delineation of anatomical structures and abnormalities. The results of our study provide valuable insights into the strengths and weaknesses of each model, offering important considerations for researchers and practitioners in the field of medical image analysis.

Our findings demonstrate that the SAM model consistently outperformed the other two models, HQ-SAM and YOLOv8, in most scenarios. With its higher mean Dice Score, Precision, Recall, and F1-score, the SAM model showcased superior segmentation accuracy and overall performance. This can be attributed to the model's advanced architecture, incorporating convolutional neural networks and attention mechanisms, which effectively learn and represent complex patterns in medical images.

Our study emphasizes the importance of carefully selecting the appropriate model based on the specific requirements of the medical imaging task and the characteristics of the dataset. While the SAM model excelled in both X-ray and lung CT scan segmentation datasets, its performance on the Short-axis Aorta Ultrasound dataset was slightly lower. This underlines the need to evaluate models on diverse datasets to gain a comprehensive understanding of their capabilities.

Our research also highlights the potential for further improvement in medical image segmentation models. Future studies could focus on enhancing the SAM model or exploring new architectures and attention mechanisms to advance segmentation accuracy. Additionally, investigating model ensembles or combining the strengths of multiple models could lead to even higher segmentation performance.

In conclusion, our study underscores the effectiveness of the SAM model for medical image segmentation tasks, particularly in X-ray and lung CT scan datasets. While HQ-SAM offers potential advantages, careful consideration of its computational cost is essential. YOLOv8, while lagging behind in performance, requires further refinement to become a viable option for medical image segmentation.